\documentclass[runningheads]{llncs}
\usepackage{mathtools}
\usepackage{graphicx}
\usepackage{cite}
\usepackage{makecell}
\usepackage{multirow}
\usepackage{hyperref}
\usepackage{xcolor}
\usepackage{amsmath}
\usepackage{amssymb}
\usepackage{multirow}
\usepackage{marvosym}
\usepackage[ruled,vlined]{algorithm2e}

\definecolor{commentcolor}{RGB}{110,154,155}   
\newcommand{\PyComment}[1]{\ttfamily\textcolor{commentcolor}{\# #1}}  
\newcommand{\PyCode}[1]{\ttfamily\textcolor{black}{#1}} 
\newcommand*\samethanks[1][\value{footnote}]{\footnotemark[#1]}
\begin{document}
\title{Detecting noisy labels with repeated cross-validations}
\titlerunning{Detecting noisy labels with repeated cross-validations}
\author{Jianan Chen 
\inst{1,2,\thanks{Equal contributions, alphabetical order} }\Letter \and Vishwesh Ramanathan\inst{1,2,\samethanks[1]}  \and Tony Xu\inst{1,2} \and  
Anne L. Martel\inst{1,2} \Letter
}

\authorrunning{J. Chen et al.}

\institute{
Department of Medical Biophysics, University of Toronto, Toronto, ON, CA  
\\
\email{chenjn2010@gmail.com; a.martel@utoronto.ca} \and
Sunnybrook Research Institute, Toronto, ON, CA}
\maketitle              

\begin{abstract}
Machine learning models experience deteriorated performance when trained in the presence of noisy labels. This is particularly problematic for medical tasks, such as survival prediction, which typically face high label noise complexity with few clear-cut solutions. Inspired by the large fluctuations across folds in the cross-validation performance of survival analyses, we design Monte-Carlo experiments to show that such fluctuation could be caused by label noise. We propose two novel and straightforward label noise detection algorithms that effectively identify noisy examples by pinpointing the samples that more frequently contribute to inferior cross-validation results. We first introduce Repeated Cross-Validation (ReCoV), a parameter-free label noise detection algorithm that is robust to model choice. We further develop fastReCoV, a less robust but more tractable and efficient variant of ReCoV suitable for deep learning applications. Through extensive experiments, we show that ReCoV and fastReCoV achieve state-of-the-art label noise detection performance in a wide range of modalities, models and tasks, including survival analysis, which has yet to be addressed in the literature. Our code and data are publicly available at \url{https://github.com/GJiananChen/ReCoV}.

\end{abstract}
\section{Introduction}
Label noise, where labels inaccurately represent the data, is ubiquitous in real-world machine learning datasets. Training machine learning models in the presence of noisy labels may lead to deteriorated performance and inaccurate conclusions. Mislabelling can arise from various sources such as human error, inadequate data quality, and encoding errors \cite{frenay2013classification}. In medical data, attaining a definitive gold standard label is often intractable, and is further complicated by inter-observer variability. Consequently, label noise in medical datasets is in general more complex and more difficult to address \cite{karimi2020deep, ju2022improving}.

A substantial body of research has been dedicated to the problem of label noise in classification tasks. A major branch of such methods involves detecting and removing noisy labels based on the metrics, predictions and uncertainties produced by the classification model itself. However, the performance of these approaches is then limited by model robustness, and may suffer from over- or under-detection of noisy labels \cite{frenay2013classification}. Another avenue focuses on increasing model reliability despite the known presence of label noise, with techniques such as robust loss functions and network structures, semi-supervised data selection and data re-weighting \cite{matuszewski2018minimal, xu2021noisy,ijcai2023p494}. These algorithms have also shown promise but face the risk of over-fitting to specific datasets. In our work, we choose to focus on the explicit detection of noisy labels as it offers an opportunity for experts to review and potentially re-evaluate ambiguous and challenging cases \cite{matic1992computer}.

\begin{figure}
    \centering
    \includegraphics[width=\textwidth]{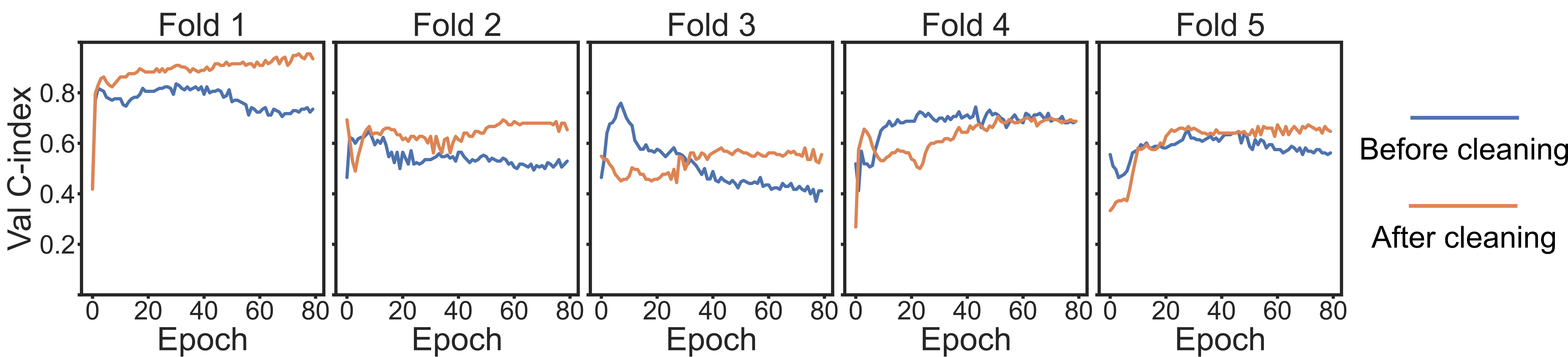}
    \caption{Validation performance (concordance-index) on the TCIA Head-Neck-PET-CT dataset, comparing performance pre- and post-cleaning of noisy samples.} 
    \label{fig1}
\end{figure}

The differences in model performance across cross-validation folds are usually perceived as completely random or as ``normal'' fluctuations caused by different seeds and software implementations \cite{krogh1994neural,reinke2024understanding}. Nevertheless, inspired by the frequent observation of large performance gaps between different cross-validation folds in outcome prediction tasks, we hypothesize that such differences may not be entirely random but could be influenced by label noise, and if so, such influence may be leveraged for identifying noisy samples (\textbf{Fig. \ref{fig1}}). We thereby propose Repeated Cross-Validations (ReCoV), a model-agnostic and parameter-free workflow for label noise detection, and fastReCoV, a less powerful but significantly more efficient variant of ReCoV tailored to resource-intensive modern deep learning frameworks. With extensive experiments in a large variety of datasets, models, and tasks, we empirically corroborate the influence of label noise on fold-specific validation performance and demonstrate that ReCoV and fastReCoV achieve state-of-the-art performance in noisy label detection in medical imaging datasets with real-world label noise.
\section{Methods}

\subsubsection{Assumptions}
Similar to existing research in this field \cite{goldberger2016training, northcutt2021confident}, we assume that a class-conditional noise process maps the true label $y$ to the observed label $\tilde{y}$. Specifically, for each label index $j$, there exists an independent mislabeling probability $p(\tilde{y}=i \mid y=j)$, where $i,j$ belong to a label set $[m]=\{1,2,...,m\}$.


\subsubsection{Modeling cross-validations in classification task}

In a dataset with $N$ samples, among which $\epsilon N$ samples are noisy, where $\epsilon \in (0,0.5)$ is the noise ratio. For a $k$-fold cross-validation, the probability of each sample landing in a specific fold follows $\mathrm{Bernoulli}(\frac{1}{k})$. The expected number of noisy samples in each ordered fold (\textit{i.e.} folds ordered from the most number of noisy samples to the least number of noisy samples) follows $\mathrm{Hypergeometric}(\mathrm{N}, \epsilon \mathrm{N}, \mathrm{N}/k)$ \cite{harkness1965properties}. The expected number of noisy samples in the fold with the highest number of noisy samples $\mathbb{E}(n_\mathrm{most})$ can be calculated by finding the maximum of the probability mass function of the hyper-geometric distribution \cite{david2004order}, while the expected number of noisy samples in each fold $\mathbb{E}(n_\mathrm{mean})=\frac{\epsilon \mathrm{N}}{k}$. In \textbf{each iteration} of cross-validation with a completely random split, $\mathbb{E}(n_\mathrm{diff}) = \mathbb{E}(n_\mathrm{most})-\mathbb{E}(n_\mathrm{mean})$ \textbf{more noisy samples} are in the \textbf{fold with the highest number of noisy samples} compared to the average. It also follows that the distribution of clean and noisy samples in the fold with the most noisy samples (coined as the occurrence distributions) follow two Binomial distributions $\mathrm{B}(n_\mathrm{fold}, p_\mathrm{clean})$ and $\mathrm{B}(n_{\mathrm{fold}}, p_{\mathrm{noisy}})$, where
\begin{equation}
    n_\mathrm{fold}=\frac{\mathrm{N}}{k}
\end{equation}
\begin{equation}
    p_\mathrm{noisy}= \frac{\mathbb{E}(n_\mathrm{most})}{\mathrm{N}/{k}}
\end{equation}
\begin{equation}
    p_\mathrm{clean}=1- p_\mathrm{noisy}
\end{equation}


\subsubsection{Repeated Cross-Validations (ReCoV) for noise detection}

\begin{algorithm}[h]
\SetAlgoLined
    \PyComment{N\_{runs}: number of runs} \\
    \PyComment{k: number of folds} \\
    \bigbreak
    \PyCode{seeds = GenerateRandomNumbers(N\_{runs})} 
    \PyComment{generate N\_{runs} seeds }\\
    \PyCode{candidates = []}  \PyComment{initialize candidates as an empty list}\\ 
        \bigbreak
    \PyCode{for seed in seeds:} \PyComment{repeat for N\_{runs} of different seeds} \\
    \Indp   
        \PyCode{train\_sets, val\_sets = FoldSplit(data, k, seed)} \PyComment{k-fold split} \\ 
        \PyCode{models.train(train\_sets)} \PyComment{train k models with k train\_sets} \\
        \PyCode{val\_metrics = models.test(val\_sets)} \PyComment{evaluate trained models} \\
        \PyCode{worst\_set = val\_sets[Argmin(val\_metrics)]} \PyComment{find the worst fold} \\
        \PyCode{candidates.append(worst\_set.ids)} \PyComment{add `worst' ids to candidates}      \\  
    \Indm 
        \bigbreak
    \PyComment{calculate number of occurrences for each sample}\\
    \PyCode{samples, counts = Unique(candidates)}
\caption{Pseudocode of ReCoV in a Python-like style}
\label{algo:recov}
\end{algorithm}

We propose ReCoV (\textbf{Algorithm \ref{algo:recov}}) to identify noisy examples by pinpointing the samples that more frequently contribute to inferior cross-validation results. In each repeated run of cross-validations with a different data split, the indices of samples in the worst-performing fold are appended to a list, referred to as the candidate pool. As more runs are recorded, the standard deviations of the occurrence distributions of clean and noisy samples increase at a lower rate compared to the increase in the difference of their means in relation to $\mathrm{N}_{runs}$. As a result, the distributions become more separated with the increase of $\mathrm{N}_{runs}$. With a large number of runs, the occurrence distributions will be clearly separable, preventing over-/under- detection. The effect of other factors such as model strength or data imbalance will be alleviated with repeating multiple independent runs.

\begin{algorithm}[h]
\SetAlgoLined
    \PyComment{N\_{runs}: number of runs} \\
    \PyComment{k: number of folds} \\
    \PyComment{$p$: a list of sampling probabilities for each sample} \\
    \PyComment{$\tau$: temperature parameter controlling sampling process} \\
    
    \bigbreak
    \PyCode{memory = Zeros(len(data))}  \PyComment{initialize memory as 0's}\\ 
    \PyCode{identified = []}  \PyComment{initialize identified noisy samples as empty}\\ 
    \PyCode{p = Ones(len(data))/len(data)}  \PyComment{initialize p to be uniform}\\ 
        \bigbreak
    \PyCode{for run in range(N\_{runs}):} \PyComment{repeat for N\_{runs} times} \\
    \Indp   
    \PyComment{split data by weighted sampling w/o replacement} \\ 
        \PyCode{train\_sets, val\_sets = WeightedSample(data, k, p, replace=False)} \\
        \PyComment{Randomly drop $\alpha$ identified noisy samples from training} \\
        \PyCode{train\_sets = DropNoisy(train\_sets, identified, $\beta$)} 
 \\ 
        \PyCode{models.train(train\_sets)} \PyComment{train k models with k train\_sets} \\
        \PyCode{sample\_val\_metric = models.test(val\_sets)} \PyComment{calculate metrics} \\
        \bigbreak
        \PyComment{Update memory with exponential moving average across runs} \\
        \PyCode{memory = $\alpha$ * memory + (1-$\alpha$) * sample\_val\_metric}  \\
        \PyCode{identified = data[memory<T])} \PyComment{identify with threshold T} \\
        \bigbreak
        \PyCode{p = Softmax(memory/$\tau$)]} \PyComment{update sampling probabilities} \\

    \Indm 
\caption{Pseudocode of fastReCoV in a Python-like style}
\label{algo:fastrecov}
\end{algorithm}

\subsubsection{FastReCoV}
In order to make ReCoV computationally feasible for deep learning models and high-dimensional inputs, we introduce fastReCoV (\textbf{Algorithm \ref{algo:fastrecov}}). Based on the framework of ReCoV, fastReCoV achieves much improved efficiency with a few modifications.

In fastReCoV, we employ a genetic-algorithm-inspired approach to solve the optimization problem of making noisy samples accumulate in the worst fold faster. Specifically, we construct a memory bank that stores the probability of each sample being noisy, where a lower memory value means a higher chance for the sample to be noisy. The memory bank is updated in each run using exponential average with a task-specific ranking metric (\textbf{Table \ref{tab:hyper}}). A weighted sampling function based on memory values is incorporated to give low memory value samples higher chance to be assigned to the worse folds. The probabilities used in the weighted sampling involves a temperature parameter $\tau$ that balances the trade-off between efficiency and the potential to find better local minimums. Bigger $\tau$ induce a more uniform distribution and relatively random splits, and smaller $\tau$ leads to a sharper distribution, creating more greedy searching of noisy samples. The memory value for each sample is also used for segregating noisy samples from clean samples. The segregation threshold can be a predefined threshold, a predefined quantile, or based on Gaussian mixture models. Additionally, we implement a procedure in which a certain percentage $\beta$ of the most noisy samples are randomly excluded from the training process in each run, to improve the robustness of trained models.

\begin{table}[]
\caption{Hyperparameters used in fastReCoV experiments. Thresholds are either absolute values or percentiles. For CIFAR-10N we choose probability of the image belonging to the given ground true label as the ranking metric. For HECKTOR, we created our own ranking metric inspired from c-index. For a particular sample, we evaluated its concordance with all the other samples both within and across the folds. For PANDA, we used the absolute distance between the ground truth label and predicted label. }
\begin{tabular}{cccc}
\hline
\textbf{Dataset}    & \textbf{CIFAR-10N}    & \textbf{HECKTOR}               & \textbf{PANDA}               \\ \hline
Sample-level metric & predicted probability & sample-level concordance  &  regression distance \\
Threshold $T$         & 0.3                   & 4\%                            & 10\%                          \\
N\_runs             & 10                    & 50                             & 15                           \\
Temperature $\tau$     & 0.1                   & 0.5                            & 1.0                          \\
Drop rate $\beta$     & 0.8                   & 0.1                            & 0.5                          \\
EMA weight $\alpha$    & 0.3                   & 0.3                            & 0.3                          \\ \hline 
\label{tab:hyper}
\end{tabular}
\end{table}

\section{Experiments}
We performed experiments in four public datasets with various tasks, models, and types of noises. Specifically, we used Mushroom and CIFAR10N as sandboxes to lay out mathematical foundations and quantitatively evaluate noisy label detection performance. We then included PANDA and HECKTOR to ensure that our method is practical and robust in large-scale real-world medical imaging datasets. 5-fold cross-validation is used in all ReCoV experiments. For detailed descriptions about model structures please refer to the corresponding references and \href{https://github.com/GJiananChen/ReCoV}{our Github repository}. Hyperparameters of fastReCoV for different experiments are summarized in \textbf{Tabel S1}. 


\subsubsection{Binary classification on Mushroom with ReCoV}

The Mushroom dataset (n=8124, \url{https://archive.ics.uci.edu/ml/datasets/mushroom}) is a widely used noisy label detection benchmark from the UCI data repository. Mushroom comprises 22 categorical features (converted to 117 dummy variables) used to predict whether a mushroom is poisonous). \\
\textbf{Experimental design:} First, we simulate class-conditional label noise by randomly flipping a proportion of labels ($\epsilon=0.1$). Next, we run a Monte-Carlo simulation to simulate how  
clean and noisy samples are sampled in the most noisy fold (fold with the highest number of noisy samples) and accumulate in the candidate pool across runs. The Monte-Carlo simulation only considers random sampling in cross-validations and does not take model training or evaluation into account. We then run ReCoV on the same noisy dataset for the same number of runs as the Monte-Carlo simulation, recording samples in the fold with the worst validation performance to generate experimental occurrence distributions. The purpose of this experiment is to evaluate the similarities between the theoretical and experimental occurrence distributions, where a high match would suggest that the fold with the highest number of noisy samples is most often the fold with the worst validation performance, thereby suggesting an association between label noise and fold-specific validation performance. 

A logistic regression model is trained to perform binary classification. We compare ReCoV with two state-of-the-art algorithms, namely Confident Learning \cite{northcutt2021confident}, and Clustering TRaining Losses (CTRL) \cite{yue2022ctrl} by comparing their accuracy in noise detection, as well as the accuracy of the retrained models after removing identified noisy samples. 

\subsubsection{Multi-class classification on CIFAR-10N with fastReCoV}

The CIFAR-10 dataset is a classic computer vision dataset that consists of 50000 training images and 10000 test images from 10 classes \cite{krizhevsky2009learning}. The CIFAR-10N dataset is a variant of CIFAR-10 with multiple sets of human-annotated real-world label noise introduced by crowd-sourcing annotations from Amazon Mechanical Turk \cite{wei2022learning}. We selected the ``aggre'' (9.03\% noise ratio) and ``worst'' (40.21\% noise ratio) label sets to reflect different noise level settings. This dataset is used to investigate the performance and behavior of fastReCoV in a modest-sized dataset with real-world label noise. \\
\textbf{Experimental design:} We compared two different feature extractors: a ViT-base-patch16 model pretrained on ImageNet-21k in a supervised fashion ($n_{dim}=768$) \cite{dosovitskiy2020image}, and a ViT-S/14 model pretrained on 142 million images with DINOv2 in a self-supervised fashion ($n_{dim}=384$, stronger features) \cite{oquab2023dinov2}. A logistic regression is trained based on the extracted features to predict the class of images. We evaluate our method with the accuracy of the retrained model after removing identified noisy samples.

\subsubsection{Survival prediction on HECKTOR with ReCoV and fastReCoV}

For noise detection in the medical domain, we first evaluate fastReCoV on the training set of HECKTOR2022 (the test set was not released), which contains 524 (390 for training and 134 for held-out test) PET-CT images of head and neck cancer patients from multiple institutes for a survival prediction challenge\cite{andrearczyk2022overview}. Survival prediction datasets are naturally noisy, especially with noise from additional sources such as loss of follow-up and disease-irrelevant mortality. In this experiment, we evaluate the ability of ReCoV and fastReCoV for detecting noisy samples in outcome prediction, a relatively unexplored field in noise detection due to the difficulty in dealing with the time-to-event labels. We evaluate our method with the concordance-index of the retrained model after removing identified noisy samples. \\
\textbf{Experimental design:} Radiomic features of the CT images are extracted using pyradiomics based on the provided individual tumor segmentations \cite{van2017computational}. A multiple instance survival neural network is trained to predict 2-year survival outcomes of patients based on multifocal tumor features \cite{chen2021aminn, chen2022metastatic}. 4000 runs of ReCoV is performed.

\subsubsection{Regression on PANDA with fastReCoV}

The Prostate Cancer Grade Assessment (PANDA) dataset, is a challenge dataset focused on predicting ISUP grade 1-5 (or 0 for benign) from Hematoxylin \& Eosin stained whole-slide images of prostate biopsies \cite{bulten2022artificial}. The dataset comprises 11,000 whole-slide images for training from Radboud University Medical Center and Kariolinsa Institute. Two external datasets with 844 and 4,675 images, respectively, are available for evaluation. The noise for this dataset mostly comes from different levels of annotator expertise in different centers, and the inter-observer variability for a subjective grading task. \\
\textbf{Experimental design:}
For the image regression task on the PANDA dataset, we first extract patches at a regular stride, at 1 $\mu m/\mathrm{px}$, and derived feature vectors using the pretrained CTransPath model \cite{wang2022}. These feature vectors were then combined using Trans-MIL \cite{shao2021transmil}, a multiple instance learning classification model. The model was trained using cross-entropy of the predicted probabilities and the annotated grade. We compare fastReCoV with the noise detection strategy of the winning team for the PANDA challenge, and their approach is to remove samples based on the gap between the prediction of their model and the ground truth. The approaches are evaluated using the quadratic weighted kappa metric of the retrained models after removing identified noisy samples.

\section{Results}
\begin{table}[]
\centering
\caption{Quantitative comparison of ReCoV, Confident learning (CL) and Clustering TRaining Losses for label error detection (CTRL) in three repeated trials, accuracy values are presented in the format of mean $\pm$ standard deviation. For performance of CL and CTRL, we took results reported in the CTRL paper.}
\begin{tabular}{|c|c|c|c|}
\hline
\textbf{Noise ratio} & \textbf{Method} & \textbf{Mask accuracy} & \textbf{Retrained model accuracy} \\ \hline
\multirow{4}{*}{10}  & No clean        & 90.0 $\pm$ 0.0               & 94.7 $\pm$ 0.8                          \\
                     & CL          & 97.1 $\pm$ 0.3               & 99.3 $\pm$ 0.3                          \\
                     & CTRL           & 99.9 $\pm$ 0.1               & 99.9 $\pm$ 0.1                          \\
                     & \textbf{ReCoV}  & \textbf{100} $\pm$ \textbf{0.0}       & \textbf{100} $\pm$ \textbf{0.0}                  \\ \hline
\multirow{4}{*}{20}  & No clean        & 80 $\pm$ 0.0                 & 86.1 $\pm$ 1.3                          \\
                     & CL              & 89.2 $\pm$ 0.7               & 95.5 $\pm$ 0.7                          \\
                     & CTRL            & 98.9 $\pm$ 0.2               & 99.2 $\pm$ 0.2                          \\
                     & \textbf{ReCoV}  & \textbf{100} $\pm$ \textbf{0.0}       & \textbf{100} $\pm$ \textbf{0.0}  \\  \hline             
\end{tabular}

\label{tab:recov}
\end{table}
\begin{table}[]
\centering
\caption{Performance of label noise detection algorithms on CIFAR-10N. Accuracy of retrained models on the test set are reported.}
\begin{tabular*}{\linewidth}{@{\extracolsep{\fill}} cccccc }
\hline
\multicolumn{1}{c|}{\textbf{Features}} & \textbf{Clean} & \multicolumn{1}{c|}{\textbf{Noisy}} & \textbf{fastReCoV}          & \textbf{Naive} & \textbf{Random}                       \\ \hline
\multicolumn{6}{c}{\textit{CIFAR-10N-aggre, Noise Ratio=9.03\%}}                                                                              \\ \hline
\multicolumn{1}{c|}{ImageNet} & 94.54 & \multicolumn{1}{c|}{92.97} & \textbf{94.84} & 94.33  & {\color[HTML]{FE0000} 92.51} \\
\multicolumn{1}{c|}{DINOv2}   & 96.46 & \multicolumn{1}{c|}{95.23} & \textbf{96.63} & 96.19  & {\color[HTML]{FE0000} 95.07} \\ \hline
\multicolumn{6}{c}{\textit{CIFAR-10N-worst, Noise Ratio=40.21\%}}                                                                             \\ \hline
\multicolumn{1}{c|}{ImageNet} & 94.54 & \multicolumn{1}{c|}{86.69} & \textbf{89.58} & 88.99  & {\color[HTML]{FE0000} 86.46} \\
\multicolumn{1}{c|}{DINOv2}   & 96.46 & \multicolumn{1}{c|}{90.26} & \textbf{92.25} & 90.58  & {\color[HTML]{FE0000} 90.02} \\ \hline
\end{tabular*}
\label{cifar10N}
\end{table}
The occurrence distributions from running ReCoV match the Monte Carlo simulations strikingly well (\textbf{Fig. \ref{fig1}}). Our results on \textbf{Mushroom} empirically corroborate the association between cross-validation fold-specific validation performance with the prevalence of label noise in the folds. In fact, as the distributions are highly similar, it becomes straightforward to calculate the number of runs required to achieving any given percentage of overlap between noisy and clean sample distributions. In (\textbf{Fig. \ref{fig1}}) we show results for 4.5\% (2$\sigma$), 0.3\% (3$\sigma$) and $\sim$0\% overlap of noisy and clean sample distributions. This result suggests that with a large number of runs ReCoV can accurately separate clean and noisy samples without the need for finding a separation threshold. When compared to existing noise detection methods, ReCoV outperforms Confident learning (CL) \cite{northcutt2021confident} and Clustering TRaining Losses \cite{yue2022ctrl} for label error detection (CTRL) by achieving a perfect separation between clean and noisy samples and perfect retrained model accuracy in both 10\% and 20\% noise ratio (p$<$0.001, DeLong's test, \textbf{Table \ref{tab:recov}}).

\begin{figure}
    \centering
    \includegraphics[width=10cm]{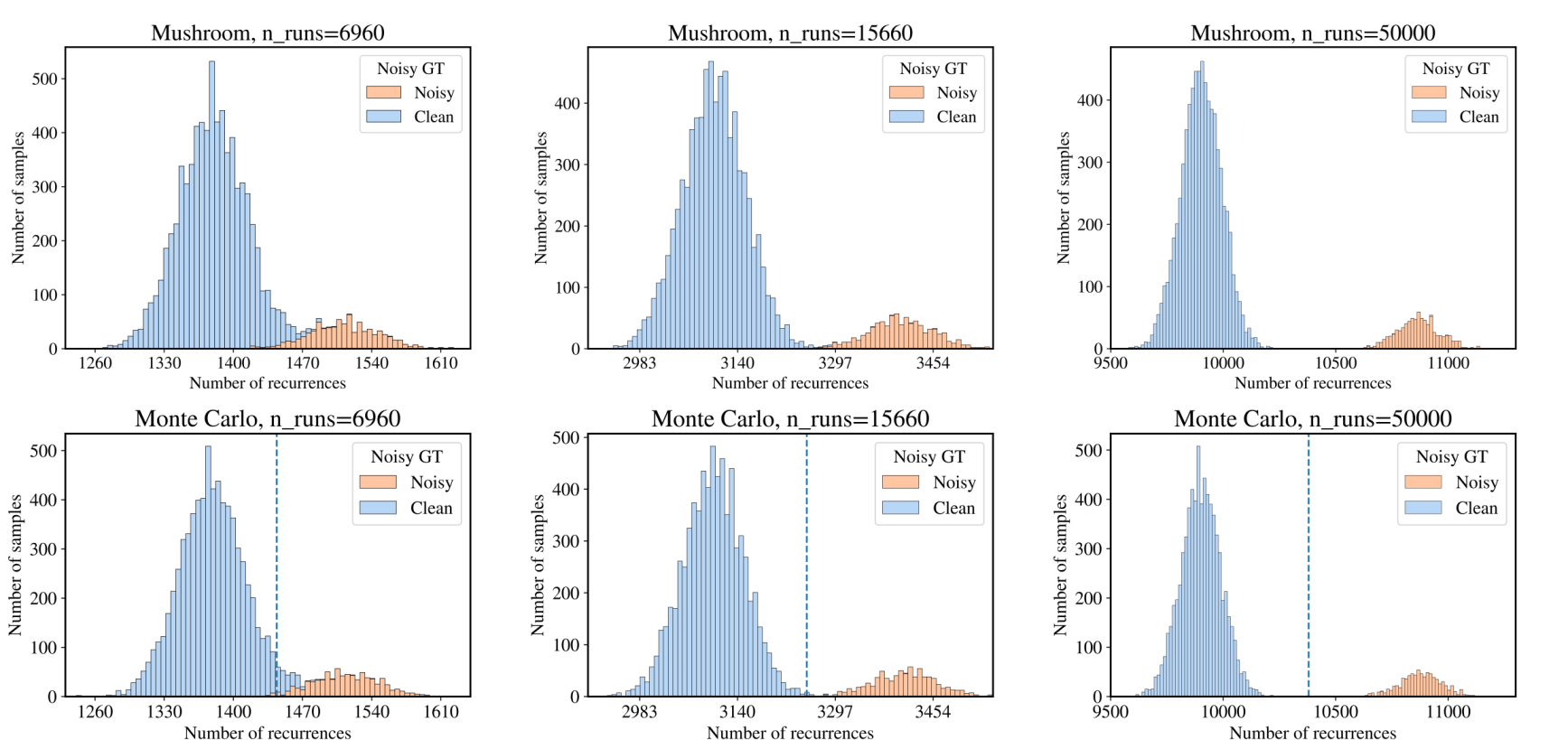}
    \caption{Results on the Mushroom dataset (N=8124, noise ratio=10\%) matches Monte Carlo simulations. Number of runs are selected to show 4.5\%, 0.3\% and $\sim$0 overlap of noisy and clean sample distribution. Dashed lines in the Monte Carlo plots refer to the theoretical separation thresholds.} 
    \label{fig1}
\end{figure}

In \textbf{CIFAR-10N}, fastReCoV consistently achieves the best noise detection performance with both sets of features and both noise ratios (\textbf{Table \ref{cifar10N}}). Two approaches are compared with fastReCoV. ``Naive'' stands for the process of identifying noisy cases according to the disagreement between model predictions and ground truth. Though it is a simple method, Naive is easy to implement, and relatively generalizable. Naive is often the first choice when tackling complex medical data and was the approach adopted by the top performing team for the PANDA challenge \cite{bulten2022artificial}. Random noise detection, which means randomly removing samples from the training set, serves the purpose of an ablation test.  FastReCoV detected the real-world annotation errors with sensitivity of 93\% and specificity of 99\% for the ``aggre'' noise level and sensitivity of 92\% and specificity of 98\% for the ``worst'' noise level.

The experiments in CIFAR-10N highlights the robustness of fastReCoV for detecting real-world annotation noise with different noise levels and models of different strengths. It's worth noting that the retrained models with fastReCoV achieved better performance compared to training on clean data in the ``aggre'' noise setting. Since the feature extractors were frozen during training and inference, this result suggests that fastReCoV removed some hidden noisy samples or confusing samples in the training set and improved model generalizability. 


To the best of our knowledge, we have proposed the first explicit label noise detection algorithm for survival analysis. In \textbf{HECTOR}, ReCoV and fastReCoV improved the concordance-index of the multiple instance survival prediction model from 0.550 to 0.635 and 0.624, respectively (\textbf{Table \ref{HECKTOR}}), which are considerable improvements in survival analysis without modifications to model structure and learning procedures. This task illustrates the trade-off between ReCoV and fastReCov: ReCoV required much higher runtime, but produced superior performance in the end.

In \textbf{PANDA}, fastReCoV again achieved the best retrained model QWK score in both held-out test sets, while Naive failed to improve model performance in test set 2  (\textbf{Table \ref{PANDA}}). A few grade 5 and benign samples with the lowest weights (\textit{i.e.} most noisy) are displayed in \textbf{Fig. S2} for further examinations. 

FastReCoV (all required runs) took 3.5mins for CIFAR10N, 3.5hrs for HECKTOR and 40.6hrs for PANDA on one Nvidia Titan Xp GPU. The algorithms are fold-independent and can be further accelerated by parallelizing multiple GPUs.
 


\begin{table}[]
\caption{Performance of label noise detection algorithms on HECKTOR. Concordance-index and standard deviation of retrained models in 100 repeats are reported.}
\begin{tabular*}{\linewidth}{@{\extracolsep{\fill}} lccccc }
\hline
 HECKTOR2022          & \textbf{Baseline} &\textbf{ReCoV}  & \textbf{fastReCoV}  & \textbf{Random}                \\ \hline
Held-out test (n=134)  & 0.550 ± 0.031     & \textbf{0.635 ± 0.030}                        & 0.624 ± 0.023  & 0.560 ± 0.045 \\
\hline
\end{tabular*}
\label{HECKTOR}
\end{table}

\begin{table}[]
\caption{Performance of label noise detection algorithms on PANDA. Quadratic Weighted Kappa of retrained models are reported.}
\begin{tabular*}{\linewidth}{@{\extracolsep{\fill}} lccccc }
\hline
           & \textbf{Baseline} & \textbf{Naive}                      & \textbf{fastReCoV}      & \textbf{Random}                \\ \hline
Test set 1 (n=844) & 0.886    & 0.898                        & \textbf{0.908} & {\color[HTML]{FE0000} 0.880} \\
Test set 2 (n=4675)& 0.866    & {\color[HTML]{FE0000} 0.836} & \textbf{0.874} & {\color[HTML]{FE0000} 0.845}  \\ \hline
\end{tabular*}
\label{PANDA}
\end{table}


\section{Conclusions and Discussion}
In conclusion, we discovered that the fluctuation of validation performance across cross-validation folds contains information on the distribution of noisy labels. Through extensive experiments, we show that this information can be leveraged to detect label noise in repeated cross-validations. ReCoV and fastReCoV are powerful and plug-and-play label noise detection algorithms that can be applied to a variety of machine learning models in a variety of supervised learning tasks.
In our experiments we used pretrained models as feature extractors instead of training end-to-end deep models. With the advances in medical foundation models \cite{azizi2023robust, chen2024towards}, we believe ReCoV-based approaches can be a practical choice. 


\begin{credits}
\subsubsection{\ackname} The authors would like to thank The Natural Sciences and Engineering Research Council of Canada (NSERC) for funding.

\subsubsection{\discintname}
The authors have no competing interests to declare that are
relevant to the content of this article. 
\end{credits}
%
%
%
%
\bibliographystyle{splncs04}

\end{document}